\documentclass{article}

\usepackage[final, nonatbib]{neurips_2019}

\usepackage[utf8]{inputenc}
\usepackage[T1]{fontenc}
\usepackage{hyperref}
\usepackage{url}
\usepackage{booktabs}
\usepackage{amsfonts}
\usepackage{nicefrac}
\usepackage{microtype}
\usepackage{graphicx}
\usepackage{xcolor}
\usepackage{lipsum}
\usepackage{biblatex}
\title{CloneBot: Personalized Dialogue-Response Predictions \\
  \vspace{1em}
  \normalsize March 22, 2021
}

\author{
  Tyler Weitzman \\
  Department of Computer Science \\
  Department of Mathematics \\
  Stanford University \\ \texttt{tylerw@cs.stanford.edu} \\
  \And
  Hoon Pyo (Tim) Jeon \\
  Graduate School of Business \\
  Stanford University \\ \texttt{hpjeon@stanford.edu} \\
}

\begin{document}

\maketitle

\begin{abstract}
Our project task was to create a model that, given a speaker ID, chat history, and an utterance query, can predict the response utterance in a conversation. The model is personalized for each speaker. This task can be a useful tool for building speech bots that talk in a human-like manner in a live conversation. Further, we succeeded at using dense-vector encoding clustering to be able to retrieve relevant historical dialogue context, a useful strategy for overcoming the input limitations of neural-based models when predictions require longer-term references from the dialogue history. In this paper, we have implemented a state-of-the-art model using pre-training and fine-tuning techniques built on transformer architecture and multi-headed attention blocks for the Switchboard corpus. We also show how efficient vector clustering algorithms can be used for real-time utterance predictions that require no training and therefore work on offline and encrypted message histories.
\end{abstract}

\section{Introduction}
We introduce \texttt{CloneBot}, a set of deep-learning based algorithms for building dialogue-response models capable of adopting a real human's persona and identity in a conversation. 

Recent advances in text to speech synthesis have made it possible to clone a human speaker's voice given a sample of just a  few seconds of audio\cite{Jia-speakers}. This breakthrough in speech synthesis begs the question: What would it take to build an underlying NLG dialogue-response system to generate the textual utterances this speaker might actually say? A speaker, for this problem, is characterized by the language they speak, the phrases they commonly use, the factual knowledge they possess about themselves and about others, and their standard reactions to various common statements and questions (do they say I love you back?). The latter trait is of course dependent on whom they are speaking to, and so part of CloneBot's job is to further account for the cross-pair  interactions between speakers, or, in the case of a group conversation, to account for all speakers involved. 

This task is of importance because user-personalized NLG models are able to provide significant productivity increases to the end-users of any communication platform. Gmail's Smart Compose feature, for example, which autocompletes word suggestions during email composition, reportedly saves Gmail users from typing over 1 billion characters each week\cite{gmail}. CloneBot's primary application is not to autonomously replace a speaker, but rather, to predict entire utterances that a user could choose to send either edited or unedited. However, an autonomous conversation with CloneBot is a great tool for testing its abilities. Unlike Gmail Smart Compose, CloneBot is trained on utterances rather than on email, and, with the growing trend of both messaging and voice applications, including Clubhouse, Discord, Speechify, and Sonic.app, it is presumable that a user's end-to-end response (including voice synthesis) could be entirely automated with CloneBot when its predictions are accurate.

Building a system like CloneBot is inherently difficult because the hypothetical one hundred percent accuracy model would require being able to exactly predict how the real user would respond. However, users may respond differently depending on the time they have available, their mood, and other random factors, not to mention that the data available to CloneBot is always going to be limited. In fact, access to good data is a primary challenge in building a system like CloneBot. CloneBot requires a far vaster amount of data and knowledge on the speaker than a few seconds of audio. Moreover, CloneBot regularly needs new data in order to be able to predict utterances by the present speaker (rather than predicting what or how they may have said it a year ago). The largest such datasets available about most users is their historical, stored communication from apps like Gmail, iMessage, WhatsApp, etc. This data is incredibly sensitive, private, and, in many cases, encrypted. The best version of CloneBot would work offline on the user's device, such that there is no need to expose private data to a centralized server. Doing so is challenging due to (1) lack of access to proper machine learning GPU hardware on consumer devices (2) limited storage space (3) the question of how to integrate, or transfer, all the stored knowledge into the bot.
CloneBot improves on Gmail's Smart Compose predictions in that the personalization piece of Gmail's Compose is a basic n-gram language model which gets combined in a weighted average with their neural-based language model \cite{gmail}. While such a model is trivial to train, it does not integrate or memorize knowledge well, nor does it generally benefit from all the benefits provided by deep-learning transformer architecture and by large pre-trained models.

The current state of the art for a chat bot holding a consistent personality is TransferTransfo\cite{TransferTransfo}, a chat bot trained on Facebook's \texttt{PersonaChat} dataset where each speaker is defined by a few sentences such as "I have a dog. I have four children." CloneBot expands the challenge to personifying a chat bot based on real user data, and based on a far vaster amount of it, without relying on a manually labeled data-set like PersonaChat. Moreover, CloneBot's ideal version also has the ability to train on a user's consumer device, a feat that we explore achieving through dense-vector clustering on a CPU.

Our approach for this paper was primarily experimentation. Our aim was to experiment heavily with various datasets, various pre-trained models, and various techniques and libraries, to learn what is possible and with what efficiency. We propose several methods we've used for achieving CloneBot functionality, along with their results, pros and cons observed, as well as our ideas for future research direction and experiments.

In total, we ran experiments across eight subsets of four datasets. Of these four datasets, three were private datasets, entirely collected for this paper. The authors used their own messaging history from iMessage, KakaoTalk, and Sonic.app to collect these datasets. Additionally, three different large pre-trained models were used across these experiments (DialoGPT, KoGPT, BERT), each pre-trained on separate datasets. Two different efficient similarity search and clustering of dense vectors algorithm libraries were used (faiss, nmslib) for CloneBot solutions that did not require any GPU training.  The CloneBots built in this paper were trained and tested in both English and in Korean.

Each of the approaches had their own benefits and limitations: Some for example, required GPU-training and a preset dataset of speaker conversations, while others relied on inference and on dense vector clustering \textit{alone}. 

\section{Related Works}
DialoGPT \cite{DialoGPT} is a large-scale pretrained dialogue response generation model for multiturn conversations. It's built on the same architecture as GPT-2, using transformer architecture with multi-headed attention blocks. It's trained on 147M Reddit conversations, so its domain is public Reddit chats. It is quite relevant for us because unlike traditional GPT, in order to capture the nature of conversations, it is trained specifically on pairs or sessions of statements: "the joint distribution of P(Target, Source)" \cite{DialoGPT}, using end of string tokens to separate each statement.  Therefore, it is the primary model we chose to fine-tune for most of our experiments.

Meanwhile, the EmpTransfo \cite{EmpTransfo} and TransferTransfo \cite{TransferTransfo} papers explore methods for creating more personable dialogue systems, which is highly relevant. Both use multi-headed outputs, with one head for  language modeling and another for multi-classification. EmpTransfo's classifier predicts emotions associated with messages, and TransferTransfo predicts which among 17 possible candidate responses is the gold truth reply. Furthermore, in EmpTransfo input embeddings are created for each category of words, emotions, actions, and these embedded vectors are added together to form the input. Likewise in TransferTransfo, input embeddings are created to represent "dialog state embedding" of which speaker is talking, which we think is the most relevant for this paper as we are trying to better distinguish and personalize between speakers.
We did adopt the input layer embedding architecture in this paper, however we were limited in our ability to adopt the multi-headed approach, however, since this technique requires supervised labeled features for the inputs of either emotion or distractors, as described in these papers, and we used our own private messaging data rather than pre-labeled sets.

TransferTransfo \cite{TransferTransfo} \cite{Golov} introduced the state of the art model for training on the PersonaChat dataset using their dialog state emeddings, and we had high hopes for this technique in distinguishing between more than two speakers. Also, the primary finding in TransferTransfo was that GPT works quite well for dialogue-response using fine-tuning, and we are excited to be one of the first to use this model to fine-tune the Switchboard corpus, detailed later in the paper. 
\section{Approach}
Our approaches can be divided roughly into three categories below.

\subsection{Fine-tuning less data between fewer speakers}
Our initial experiments approached the problem by fine-tuning DialoGPT/KoGPT for English and Korean respectively on input sequences of utterances, personalizing for the speaker solely based on the data exposed during training. For example, by exposing DialoGPT only to a conversation between Tyler and his brother, we are able to fine-tune the model to predict dialogue between these two speakers quite well.

Since DialoGPT is predicting every token based on all previously generated tokens, we condition it by teacher forcing, starting predictions with a prefix input that provides the context for what it is predicting. Our prefixed input is formatted as follows: $\{s_n, <EOS>, ..., s_7, <EOS>, s_6, <EOS>, ..., s_1, <EOS>\}$ where $s_i$ is the speech utterance spoken in the conversation $i$ turns ago, and we are next predicting $s_0$. EOS is an end-of-string token. Every $s_i$ is further encoded into subwords by our tokenizer so that $s_1 = \{w_1, w_2, ..., w_{|subwords|}\}$.  Since GPT does not simply use whole word embeddings, we can teach the model where to place punctuation as well as words, which is especially important so that a speech synthesis layer could then speak the utterance with realistic human-like pauses (as opposed to just grammatically correct pauses) if we later chose to add a speech synthesis layer following word sampler decoding. Note that the value of $n$ represents the max number of recent utterances to include in each input. It is the max number not only because some inputs begin the conversation, but also because we may not always fit as much history as we'd like depending on the length of the messages, as we are limited by the 1024 max token sequence length of DialoGPT's input transformer.

\subsection{Fine-tuning lots of data between many speakers}
In this approach, instead of fine-tuning each pair of speakers with their own separate dataset and model, we fine-tune the model on the entire set of conversations between all speakers, including even group conversations with multiple speakers, and try to help our model to learn to distinguish between each person. We performed experiments with three different such architectures: In the first architecture, we included which speaker is next to respond in the initial position of the input as an encoded string: $\{speakerID, s_n, <EOS>, ..., s_1, <EOS>\}$ . In the second approach, we included the associated speaker id for each $s_i$ before $s_i$: $$\{speakerID_n, s_n, <EOS>, speakerID_{n-1}, s_{n-1}, <EOS>, ..., speakerID_1, <EOS>\}$$And in the final approach, we expanded upon the idea introduced in TransferTransfo \cite{TransferTransfo} wherein new "dialog state embeddings" are created in the input level to be summed with the position and subword embeddings. Instead of creating just two new special tokens for <speaker1> and <speaker2> as in their source code, we instead create a new special token and speaker embedding for each individual person that appears in our dataset. We then set the inputs to $$\{<speakerID_n>, s_n, <EOS>, <speakerID_{n-1}>, s_{n-1}, <EOS>, ..., <speakerID_1>, <EOS>\}$$ 
where each $speakerID$ is now a new token with its own embedding, and we create a new token type layer identifying each input embedding to the speaker token embedding it is associated with, as shown in the figure \ref{fig:transformer} in the Appendix.

\subsection{Dense vector clustering only}
In this third category, inspired by a recent paper on Dense Passage Retrieval for Open-Domain Question Answering \cite{DensePassage}, we do not use GPT or fine-tuning techniques at all. Instead, we use BERT-based SentenceTransformers to encode vectors of past messages. Then we use these stored vectors to compute the semantic similarity between a query utterance and the sentence that is semantically the closest to the query sentence. To be specific, let $q$ be a query sentence, $BERT(q)$ the encoding output from a pre-trained BERT model, and $knn(q)$ the sentence in the chat history that is the closest in meaning to the context to $q$. Our idea is that we can think of $knn(q)$ as a reference to $q$, since we are trying to predict the next utterance prediction for $q$, we therefore simply return the associated utterance value following $knn(q)$. There are two caveats in this approach: (1) $knn(q) \neq q$ because we do not know the gold truth response yet (this is entirely unsupervised even during training). (2)  Our $knn(q, targetSpeakerID)$ method actually takes the target speaker as a parameter. Since we are trying to clone $targetSpeakerID$, we only want to return answers that have been said in the past by $targetSpeakerID$, not by somebody else. Note then that $targetSpeakerID$ is actually associated with the value that is associated with the k-nearest context (see figure \ref{fig:faiss} in the Appendix). This is our own contribution.

\section{Experiments}

\subsection{Data}
We conducted experiments with four very different datasets that address various aspects of real-world human dialogue in the wild. We describe each dataset below. Some datasets were broken down further into subsets.

\subsubsection{Switchboard Dialog Act (SwDA) Corpus}
The first dataset we used to start our experiments was the Switchboard Dialog Act (SwDA) Corpus\cite{switchboard} containing 260 hours of 2,400 two-sided telephone conversations among 543 speakers. Although this dataset was originally collected to develop audio processing technology, it provides high-quality transcripts that were put together by expert transcribers, which we leverage for utterance prediction. We used metadata provided to divide utterances between speakers in order to personalize them. We used the subsets SwDA-1047 and SwDA-1043 which were the two longest conversations in the corpus.\cite{simple_switchboard}  \ref{fig:swda} in the Appendix.

\subsubsection{Personal Chat History Data from KakaoTalk (Korean)}
The second dataset we used was Tim's personal chat history data collected on KakaoTalk to compare performances across multilingual context, in this case English and Korean. One of the challenges was determining how to properly separate out Korean text from the dialogue as the chat contained both Korean and English. The resulting dataset contains 1,152 two-sided utterances with average length of 18.6 Korean characters. One challenge we faced while working on personal chat histories was that of privacy; although there is no issue in using our own chat history for the purpose of this paper, we note that large-scale training of private data could lead to unexpected privacy issues. A sample of the dataset can be found in figure \ref{fig:kakao} in the Appendix.

\subsubsection{Personal Chat History Data from iMessage (English)}
The third dataset we used was Tyler's personal chat history collected on iMessage. Tyler extracted all 195,331 text utterances across six years of iMessage communication. Importantly, Tyler was successful to extract not only the text content, but also the metadata associated with each message including which phone number sent it, when it was sent, and what chat id it belonged to (many numbers overlapped across multiple chats due to group conversations). 

This data set was re-processed several times across various experiments with different filters and configurations. We divided it into the subsets shown in table \ref{table:imessage_sets} in the Appendix. In some cases we \textit{collapsed} repeat messages, meaning if someone sends "Hey" and then "How's it going?" for example, this collapses to the single utterance "Hey? How's it going?" We split the test set chronologically ordered by message date so as not to remove test messages in the middle of the conversation. We also ensured that every speaker in the test set was represented in the train set. 

In total there were $2,601$ phone numbers, or speakers, represented in this dataset. A sample of the dataset can be found in figure \ref{fig:imessage} in the Appendix.

\subsubsection{Personal Chat History between Tim and Tyler on Sonic}
Throughout the research of this paper the authors of this paper collaborated in communication using the Sonic app beta, founded by Tyler. We did in fact end up using our own messages to each other on Sonic to train some of the models we experimented with.


\subsection{Evaluation method}



We started with our human eye to review coherence and structure of outputs, building unit test queries that we could feed into the model automatically and inspect. We used perplexity for automated evaluation. While we used perplexity as our primary automated metric for our GPT models based on loss, we used the BLEU scoring metric as automated evaluation for our dense vector clustering bot \cite{post-2018-call}.

Finally, in order to push the bot to its limits on new human input, we built an iMessage bot that let our model directly interact with a dozen real people via iMessage. Running on Tyler's Macbook pro, a node.js listener was built to hook into the private iMessage system files to read incoming messages in real-time, parse them, send them on to a python3 instance, and finally passes the decoded logits from the model back via an apple script to reply in iMessage. Our bot kept track of as much contextual history as our model could handle and successfully referenced several messages back in ongoing chats.

\subsection{Experimental details}
\subsubsection{Fine-Tuning on Pre-Trained DialoGPT-Small}
Fine-tuning the smallest version of DialoGPT on the longest Switchboard conversation took 15-20 minutes on a single K80 GPU. Similar time was needed for other single speaker-pair conversations. Fine-tuning the larger iMessage dataset with roughly 77,000 messages took two hours per epoch on two Tesla V100 GPUs. We used an Adam optimizer with a learning rate of 5e-5, no weight decay, an Adam epsilon of 1e-8, and a block size of 512. Finally, we decode by sampling with $top_p=0.7, temperature=0.8$, which has a more human feel than beam search decoding.

\subsubsection{Fine-Tuning on Pre-Trained KoGPT2 for Korean chats}
We used KoGPT2\cite{kogpt2}, a large-scale model pre-trained on 20GB of Korean text data by SK Telecom (AT\&T of Korea). The model architecture is the same as the smallest version of GPT-2 (117MB). We adopted a similar methodology as in our fine-tuning with DialoGPT; it was interesting to see that the model pipeline is robust even for such a drastic language change from English to Korean, which is one of the most distant language pairs. We note that KoGPT2 was only partially trained on conversational data, a point which we revisit in the results section. We used the pre-trained tokenizer from the model, an Adam optimizer with a linear warmup learning rate scheduler, no weight decay, an Adam epsilon of 1e-8, and a block size of 512. We trained the model for a total of 3 epochs with a batch size of 4. For sampling, we use: $top_k=70, top_p=0.5, temp=1.2$, which are the suggested `medium-level' parameters from Hugging Face's Conversational AI demo.

\subsubsection{Using FAISS (and NMSLIB) for Sentence Similarity}
In order to implement the dense vector clustering bot, we used BERT-based SentenceTransformers in order to map sentences to their respective sentence vectors. The dimension for the sentence embeddings was set to be 1024. For English we used `paraphrase-distilroberta-base-v1' For Korean language we used `xlm-r-large-en-ko-nli-ststb' which achieved Korean semantic textual similarity (STS) SOTA benchmark of 84.05. After applying this mapping, we used FAISS for fast vector computation to get distances. We also tried NMSLIB library and found that it also could work well for our purposes. The two distance metrics used were L2-norm and cosine similarity using dot product (which requires vector scaling to get norm of 1).

\subsection{Results} 
\subsubsection{Model-Generated Text}

We first present some examples of the machine-generated text that our models outputted for each of the datasets we experimented with. We show interactions for the first 5-6 conversations where we mix up both questions and statements to feed to each model and examine their robustness. Please refer to figures \ref{fig:imessage}, \ref{fig:kakao-generated} and \ref{fig:swda-generated} in the Appendix.

\subsubsection{Evaluation: Human-Centric Metrics}
Both English and Korean speakers who reviewed output texts agreed that although not perfect, these messages were highly credible to pass as written by actual human beings. Since Tyler's iMessage bot was always responding, we got feedback like "I don't know if this is you or the bot now" or "Thanks Tyler Bot :)" or, a few times, they didn't realize it was a bot at all until we told them. The models did surprisingly well in capturing the context of the conversation and in adopting the style of speech used by the speakers in the dataset.

There were a few experiments that failed, however. When Tyler started using GPT CloneBot for his friends to talk to themselves (i.e setting the targetSpeakerID to them rather than to Tyler), the bot gave answers that were clearly not representative of how they talked. While the FAISS-based bot correctly gave answers that were only associated with the targetSpeaker, it was clear that the GPT architecture used was not successful at this task. First, most messages regardless of the targetSpeaker were over-representative of Tyler, who was the majority, over-represented speaker in the dataset. Second, in the model that had a special token layer for each speaker, while Tyler clearly did start becoming less represented, the model would still sometimes output things like "Your verification code for TikTok is 138834" even when predicting a message that was supposed to come from someone other than TikTok. However, it was quite clear that the model not only memorized quite a bit, but it did learn the structure of a conversation: Even when we did not prefix inputs for generation with a particular speaker, it would gravitate towards picking 2 real phone numbers and switching back and forth between utterances from each of them, adding the prefix for their speakerID tokens by itself.

The GPT models fine-tuned on a small dataset were quite successful, in comparison, to mimick particular speakers, because they had majority representation of those speakers.

\subsubsection{Evaluation: Automatic Metrics}
We present a summary table of perplexity scores achieved by different models that we experimented with. We show scores for two subsets of the SwDA corpus (namely SwDA-1043 and SwDA-1074, which are the two longest conversations in the corpus) and KakaoTalk. We tried using varying context lengths in order to see how model performance changes with how much context information the model is aware of. Please refer to tables \ref{table:metrics-1}, \ref{table:metrics-2}, \ref{table:metrics-3}, \ref{table:metrics-4} for detailed figures.


Our dense vector clustering bot, which was evaluated on BLEU, scored 0.37 of 1, which was a remarkable score given that this takes essentially no GPU training. It was incredibly fast. And, it was the only model that successfully sounded like any targetSpeaker we gave with just one run (as opposed to having to fine-tune a different model per speaker). We found that NMSLIB ran inference faster but FAISS can create an index faster; ultimately FAISS was faster since we had to make a separate index for each target speaker. We also created a parallel text dataset now using this bot, which could be used in a future paper to build a hybrid FAISS + GPT model.

Our test perplexity scores were quite better than expected on Switchboard conversations. With a best test perplexity of \textbf{23.05} this is a very strong score, and the model-generated utterances back it up. As far as we are aware, there has been little done with applying pre-trained models to a spoken utterance set like Switchboard, and this paper may have set a baseline to do so.

The iMessage and KaKao perplexity scores performed somewhat poorly in perplexity in comparison. We think this is because they are higher-entropy. While Switchboard participants were asked to talk with each other about a specific topic, our personal iMessage and Kakao Talk history bounce all over the place. One text we're talking about one topic and another we switch to something else. Our model did in fact learn to randomly switch topics sometimes— you would be talking with it about pizza and then it would say "I'm in the lobby," or "Just got in the Uber," or "I'm so proud of you <3" which is in fact how real humans tend to text but not at all the sort of behavior shown in Switchboard. Switchboard utterances are likely just more predictable.

\section{Analysis}
We were able to find that fine-tuning on large-scale pre-trained models like DialoGPT and KoGPT2 works surprisingly well in context-based response generation task. We discuss the qualitative findings below:

\subsection{Ability to Capture the Dialogue-Specific Characteristics}
Real-world human dialogues have their own specific characteristics: stutters, fillers, repeated words, seemingly incoherent sentences. For the models trained on SwDA dataset, we found that the trained models successfully capture these linguistic features as seen in our examples in the appendix. In addition, the models worked well in imitating a certain speaker's style of colloquial expression. We also found that the models are able to learn the features of personal styles from the chat history dataset. Another interesting observation was that despite the size differences in the pre-training and fine-tuning datasets, the fine-tuned models did not spit out random utterances using the pre-trained language model but stuck close to the conversations in the fine-tuning dataset. This hints at a welcome possibility that our CloneBot can be actually deployed in real-world products to reasonably clone the users of messaging products.

\subsection{Sensitivity to Decoding Parameters}
The text generated by these models can vary significantly depending on the parameters we chose for sampling: \texttt{top\_p}, \texttt{top\_k} and \texttt{temperature}. This is because the larger the value for \texttt{top\_k}, for example, more possible word choices are considered, meaning that the model's generation can have higher variance. Also, the \texttt{temperature} parameter controls the randomness of predictions by scaling logits that are then fed into a softmax function. By controlling this probability distribution, the \texttt{temperature} parameter can change how "conservative" the output would be. In practice, we have found that adopting a "medium-level" parameter used by Hugging Face's ConvAI demo (top\_p=0.5, temp=1.2) strikes a right balance between outputs that are too far-fetched and those that are too simple and unimaginative. Although we experimented by tweaking the dials by ourselves, we could in the future adopt a systematic approach to find the optimal parameters for decoding by optimizing for an automatic metric like BLEU. This optimization process would also have to take into account the pre-training dataset; to illustrate, our model fine-tuned on KakaoTalk data was more likely to ramble on about something that was not mentioned in the training dataset or would output a non-dialogue passage from the training dataset like a newspaper article. This is because unlike DialoGPT that is trained on conversational data, KoGPT2 was only partially trained on dialogue data. Therefore, it seems like more large-scale pre-trained datasets like DialoGPT could be very useful for further work in this direction.

\subsection{Limitations of Our Approach}
Despite our success in implementing the \texttt{CloneBot} idea using various datasets, there are clearly limitations with our approach. One of the empirical limitations that we ran into was the model's difficulty with longer chat histories. Although we imposed a maximum length of previous chat history that the model can refer to, the models would sometimes not work well when the chat histories got longer, spitting out random, unintelligible tokens. This phenomenon could be due to the length of context utterances we trained the model on, and is something we could look into more. Also we found that our model's output could be almost exactly the same as one of the utterances in the training set, as if the model simply memorized what a speaker said and returned the same expression as output. Sometimes it would be partially memorized and partially generated. This question of whether the model is memorizing or understanding the context is an important question in the research community as well, and we expect this issue to be increasingly important as the general public demands higher transparency and explain-ability of AI models. It certainly made us scared to publish checkpoints of our models, since it scarily memorized some of our private chat history.

Overall, our models fine-tuned on one conversation performed significantly better against the human evaluators— they had a more consistent personality, they seemed to respond more directly to queries—where as the models trained on 40,000 and more messages and on multiple speakers were a little noisier. They would write a whole paragraph about their planned meeting with the editor of forbes next week, or they would sign a text message "Sincerely", or they would even send us verification codes. Essentially, our approach for tokenizing different speakers missed a core feature we would add in future experiments: Penalizing the bot for saying something that someone else would have said who is not the target speaker for the prediction. It could work well to attempt a multi-headed approach. We shyed away from such an approach due to our dataset not already having built-in distractors, but we could conceivably dynamically sample distractors from other speakers in the set at random and let the model pick which one it thinks belonged to the target speaker.

\section{Conclusion}
In this paper, we explored how to implement \texttt{CloneBot} using known state of the art pre-training and fine-tuning techniques built on transformer architecture and multi-headed attention blocks. We showed that GPT can fine-tune quite well to a particular set of speakers to provide personlized outputs, and that efficient dense vector clustering is a promising alternative for being able to provide mass-scalable systems that can personalize thousands of speakers at once without the need for heavy GPU training.

We think the hybrid of the two is a promising research avenue for future work, for example, placing FAISS predictions inside of the input of a GPT model.  Our multi-pronged experimentation in this paper led us to think about many such exciting research questions for future work. One such open question is if and how we could train a bilingual conversational bot, as people often use different languages when composing messages. If the training data involves a mix of different languages, would we be able to train the model to not only recognize this fact, but also learn about the context and successfully generate in a similar multilingual style mimicking that of the user? 

Another interesting question is about how to deal with the differences in context in training and validation sets. Chat histories are often collected over a lengthy period of time and contexts in which users message each other often change over time. How could the model effectively detect and update stale information?

\printbibliography

\newpage
\appendix
\section{Appendix}
\subsection{Presentation Videos}
\begin{itemize}
    \item \href{https://drive.google.com/drive/u/0/folders/1nkuzjhIn-OO_gOwXslRl4GbGVGXMLlWW}{Video Demo 1 Link}
    \item \href{https://www.loom.com/share/ad121a9e562b4beda155187baf972df4}{Video Demo 2 Link FAISS bot demo}
\end{itemize}


\subsection{Transformer Decoder}
\begin{figure}[h]
\centering
\includegraphics[scale=0.35]{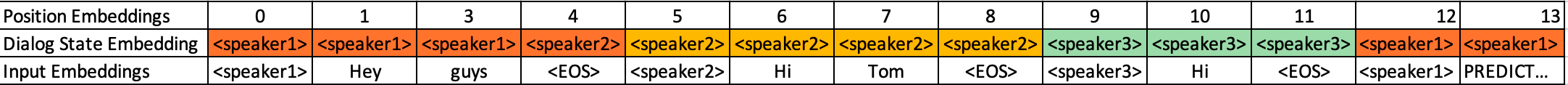}
\caption{Transformer decoder input is the sum of all three layers}
\label{fig:transformer}
\end{figure}

\subsection{FAISS/NMSLIB for KNN}
\begin{figure}[h]
\centering
\includegraphics[scale=0.4]{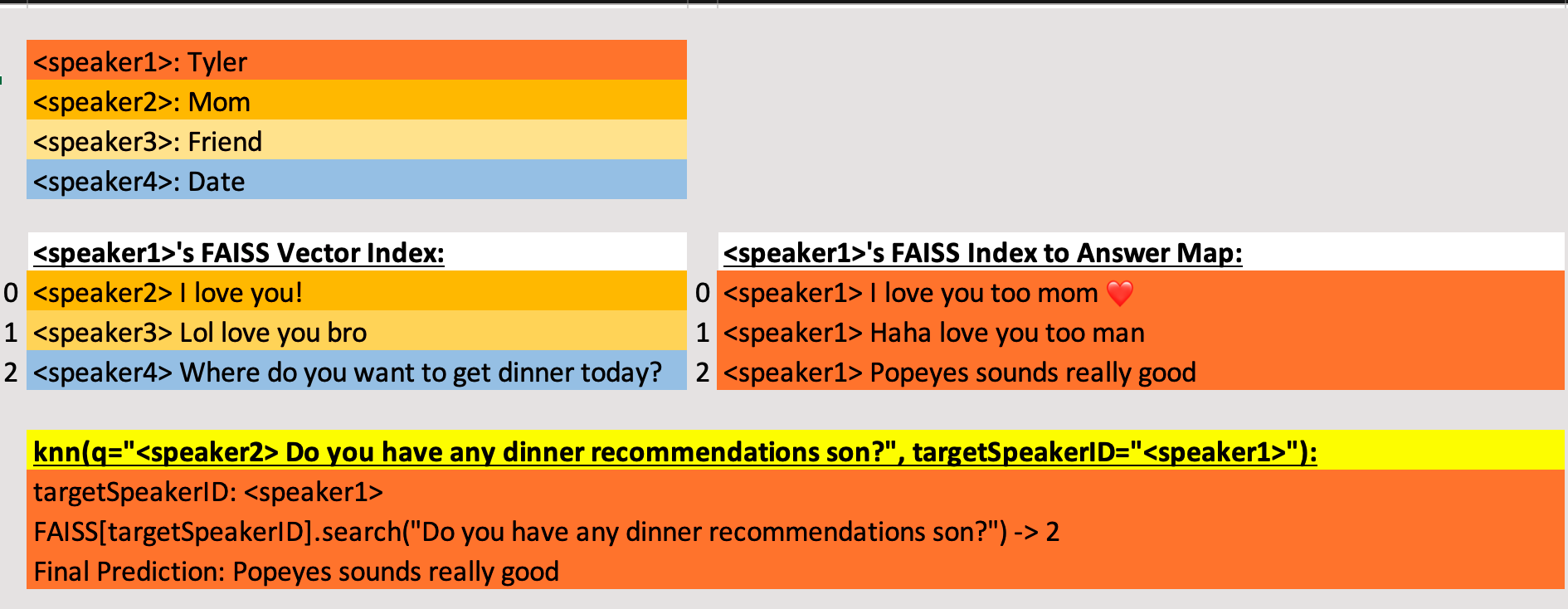}
\caption{Using FAISS/NMSLIB for KNN}
\label{fig:faiss}
\end{figure}

\subsection{Data}
\begin{figure}[h]
\includegraphics[scale=0.75]{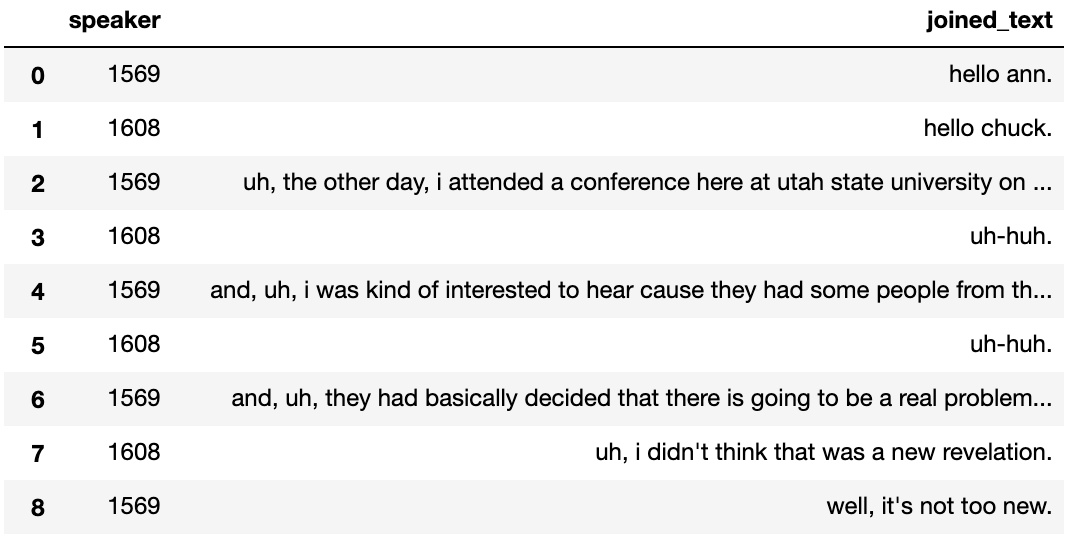}
\caption{Switchboard Dialog Act Corpus}
\label{fig:swda}
\end{figure}

\begin{figure}[h]
\centering
\includegraphics[scale=0.8]{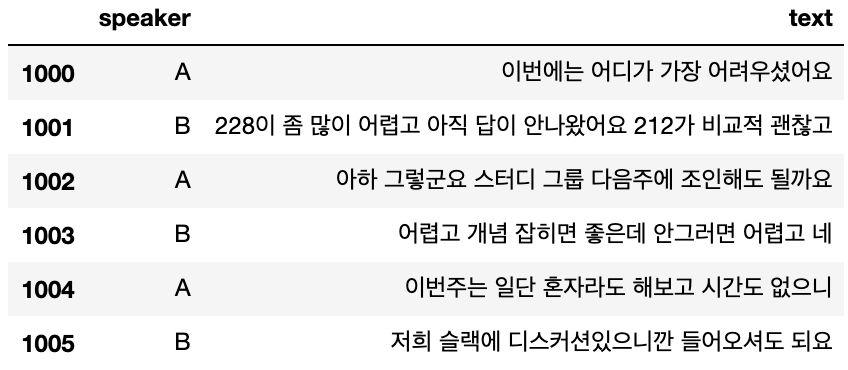}
\caption{KakaoTalk Chat History Data}
\label{fig:kakao}
\end{figure}

\begin{figure}[h]
\includegraphics[scale=0.6]{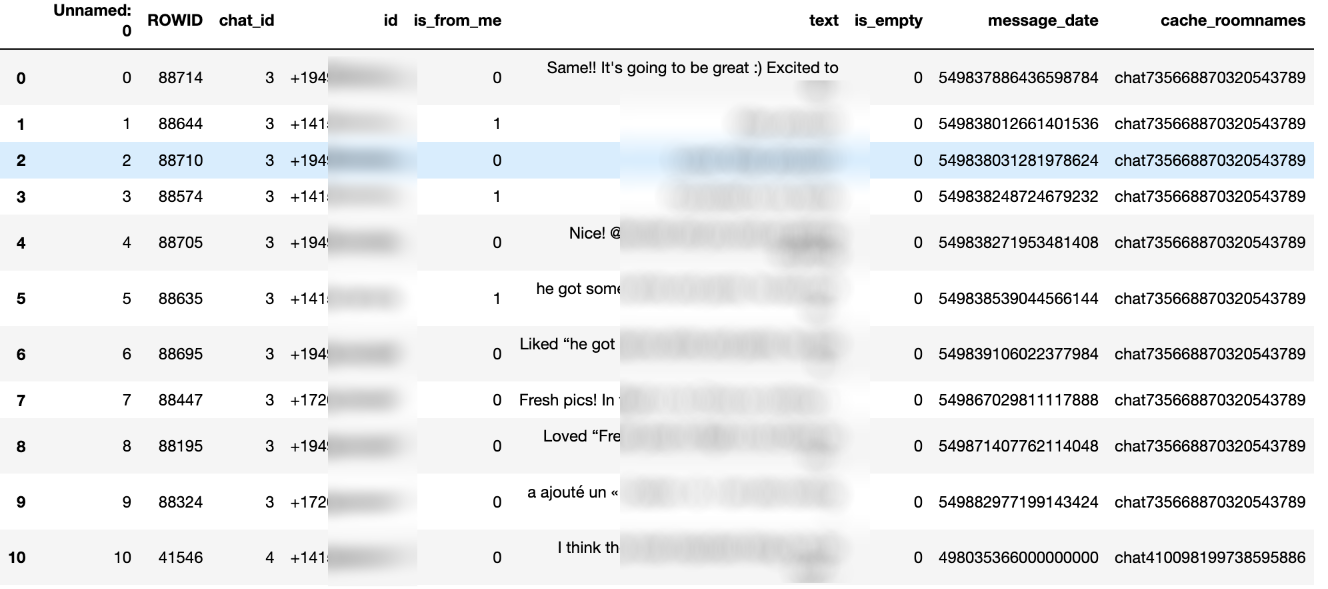}
\caption{Real iMessage Chat History Data, Blurred for Privacy}
\label{fig:imessage}
\end{figure}

\begin{table}[h!]
\begin{center}
 \begin{tabular}{|c l c|} 
 \hline
 Subset & Description & Length \\ [0.5ex] 
 \hline\hline
 iMessageAll & All Utterances & 195,331 \\
 \hline
 iMessagePairs & Pair utterances only (no group messages) & 170,488 \\ 
 \hline 
 iMessageBrother & Messages with Tyler's Brother & 9,261 \\
 \hline
 iMessageDates & Sample of "Dating" or Boyfriend-Girlfriend Messages & 3,526 \\
 \hline
\end{tabular}
\end{center}
\caption{iMessage History Subsets}
\label{table:imessage_sets}
\end{table}

\begin{table}[h!]
\begin{center}
 \begin{tabular}{|c l c|} 
 \hline
  Subset & Description &  Length \\ [0.5ex] 
 \hline\hline
 iMessage\_TRAIN & Collapsed Training Utterances & 77,973 \\
 \hline
 iMessage\_TEST & Collapsed Testing Utterances & 16,168 \\
 \hline
\end{tabular}
\end{center}
\caption{iMessage History Subsets 2}
\label{table:imessage_sets2}
\end{table}

\subsection{Model-Generated Text Samples}
\begin{figure}[h]
\includegraphics[scale=0.57]{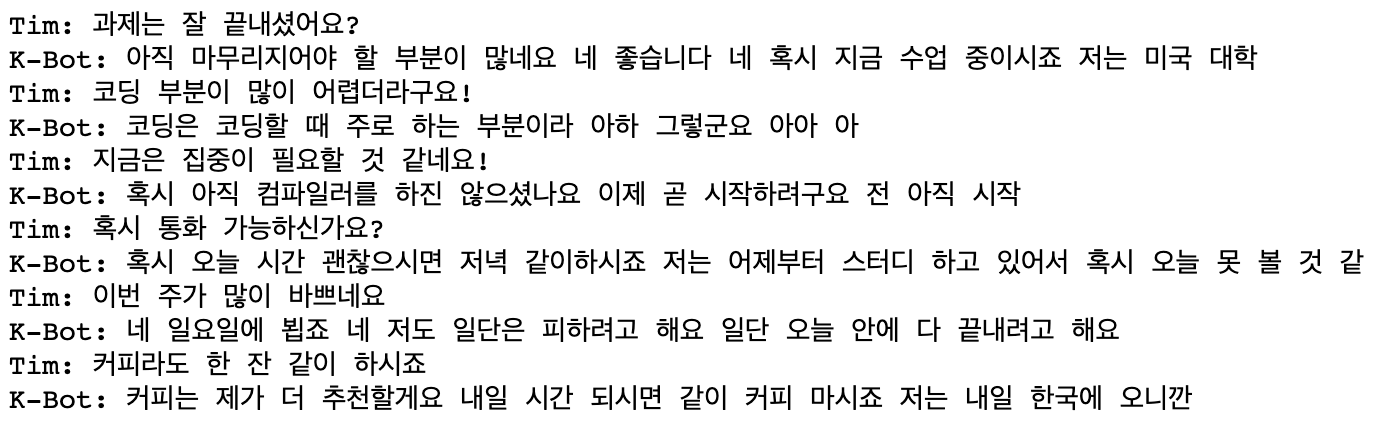}
\caption{Example of Korean Text Generated by KakaoTalk-trained Model}
\label{fig:kakao-generated}
\end{figure}

\begin{figure}[h]
\includegraphics[scale=0.4]{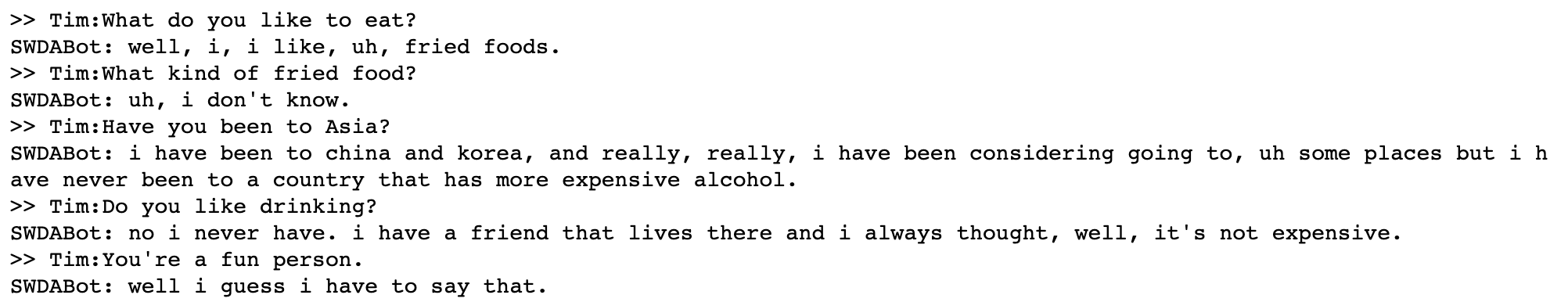}
\caption{Example of English Text Generated by SwDA-trained Model}
\label{fig:swda-generated}
\end{figure}

\begin{figure}[h]
\includegraphics[scale=0.2]{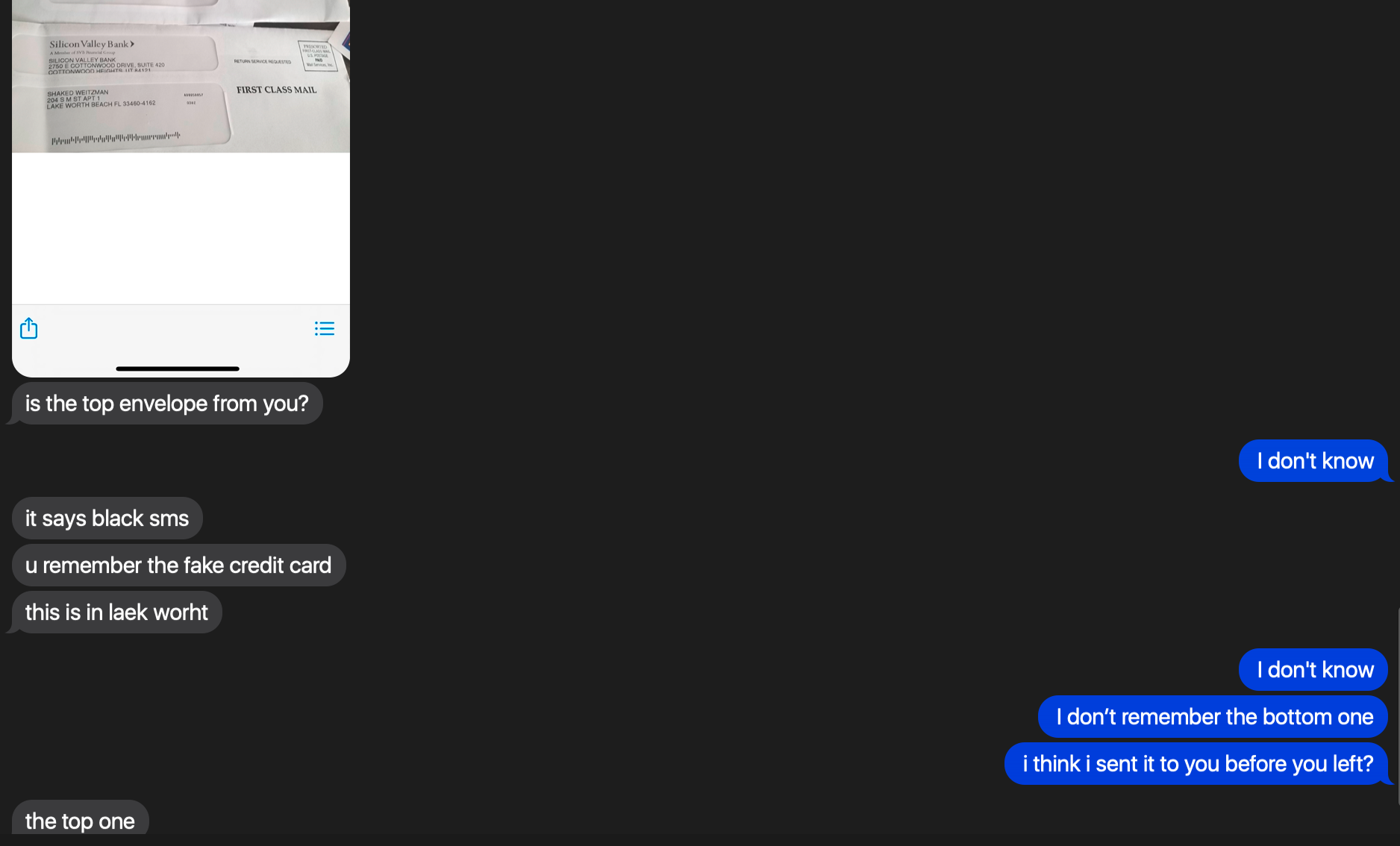}
\caption{Example of Real Conversation with iMessage-trained Model (Bot on the right, they had no idea)}
\label{fig:imessage-generated}
\end{figure}

\subsection{Metrics}
\begin{table}[h]
\begin{center}
 \begin{tabular}{|c c c c|} 
 \hline
 Context Length & SwDA-1043 & SwDA-1074 & Kakao \\ [0.5ex] 
 \hline\hline
 5 & 23.81 & 23.05 &  111.08 \\ 
 \hline
 7 & 63.12 & 30.80 & 150.52 \\
 \hline
 10 & 101.46 & 44.93 & 189.07 \\
 \hline
\end{tabular}
\end{center}
\caption{SwDA, Kakao Test Perplexity Scores}
\label{table:metrics-1}
\end{table}

\begin{table}[h]
\begin{center}
 \begin{tabular}{ | l | l | l | l | l | l | l l | }
\hline
	\textbf{ Data Trained / Data Evaluated }& Brother & Dates & All & Train & TEST \#1 & TEST \#2 & \\
	\hline
	Dates x3 (n=7) & 192.49 & 2.68* &  &  &  &  &  \\ \hline
    Dates x3 (n=15) & 214.68 &  &  &  &  &  &  \\ \hline
	Brother x3 & 6.4166* & 88.006 &  &  &  &  &  \\ \hline
	All (Unprefixed, Uncollapsed) x1 &  & 10.2373* &  &  &  &  &  \\ \hline
	All (Prefixed @0, Uncollapsed) x1 &  & 8.7556* &  &  &  &  &  \\ \hline
	All (Prefixed @0, Collapsed) x1 & 6.7576* & 3.7666* & 4.764* &  &  &  &  \\ \hline
	All (Prefixed/msg, Collapsed) x1 &  &  & 4.7484* &  &  &  &  \\ \hline
	TRAIN 40,000/77,958 messages &  &  &  & 14.7165* & \textbf{66.6131} &  &  \\ \hline
	TRAIN 77,958/77,958 messages &  &  &  & 11.4929* &  & 87.5308 &  \\ \hline
	TRAIN 77,958/77,958 x3 &  &  &  & 3.7838* & 69.4467 & 84.1496 &  \\ \hline
	 
\end{tabular}
\end{center}
\caption{Perplexity scores for DialoGPT models trained on iMessage Subsets. Table shows how the model was trained on the left and how it performed against various datasets. Values with a * are training scores. "Prefixed @0" are models in which the speaker ID was prefixed as a natural language string in the input at the first position, whereas "Prefixed/msg" prefixed the speaker ID at the beginning of each utterance. The last three rows used a multi-layer input embedding with 2,086 new special tokens (one per speaker in the dataset). "xN" represents how many repeat training epochs of the dataset occured before testing.}
\label{table:metrics-2}
\end{table}

\begin{table}[h]
\begin{center}
\begin{tabular}{ | l | l | l | l | }
\hline
	 & SwDA-1047 & Tim-Tyler & Kakao \\ \hline
	SwDA-1047 15-prefixed & 1.73* &  &  \\ \hline
	DialoGPT Switchboard 1047 7-prefixed & 2.5873* &  &  \\ \hline
	DialoGPT-small Untuned & 1188 &  &  \\ \hline
	DialoGPT-large Untuned & 1144 &  &  \\ \hline
	DialoGPT Tim-Tyler Project Chat &  & 2.18* &  \\ \hline
	KoGPT Kakao Messages (1200 messages) &  &  & 8.3155* \\ \hline
\end{tabular}
\end{center}
\caption{Other perplexity scores. Values with * are training perplexity.}
\label{table:metrics-3}
\end{table}

\begin{table}[h]
\begin{center}
 \begin{tabular}{|c c|} 
 \hline
 Model & BLEU \\ [0.5ex] 
 \hline\hline
 FAISS-only bot iMessage & \textbf{0.3747} \\
 \hline
\end{tabular}
\end{center}
\caption{FAISS-only iMessage Bot}
\label{table:metrics-4}
\end{table}

\end{document}